\documentclass[11pt,a4paper]{article}
\usepackage[hyperref]{acl2021}
\usepackage{times}
\usepackage{latexsym}
\usepackage{amssymb}
\usepackage{amsmath} 
\usepackage{booktabs}
\usepackage{enumerate}
\usepackage{graphicx}
\usepackage{subfigure}
\usepackage{xspace}
\usepackage{float}
\usepackage{bbm}
\usepackage{bm}
\usepackage{multirow}
\usepackage{booktabs}
\usepackage{color}
\usepackage{framed}
\usepackage{stfloats}
\usepackage{iitem}
\usepackage[T1]{fontenc}
\definecolor{shadecolor}{RGB}{180,180,180}
\usepackage{url}

\newcommand{\ie}{\emph{i.e.,}\xspace}

\newcommand{\eg}{\emph{e.g.,}\xspace}

\newcommand{\ignore}[1]{}
\newcommand{\tabincell}[2]{\begin{tabular}{@{}#1@{}}#2\end{tabular}}

\makeatletter
\def\@fnsymbol#1{\ensuremath{\ifcase#1\or \dagger\or *\or \ddagger\or
   \mathsection\or \mathparagraph\or \|\or **\or \dagger\dagger
   \or \ddagger\ddagger \else\@ctrerr\fi}}
\makeatother

\usepackage{microtype}

\aclfinalcopy 

\title{TextBox: A Unified, Modularized, and Extensible \\ Framework for Text Generation}

\author{
	Junyi Li\textsuperscript{\rm{1,3}}\thanks{\ \ Equal contribution.},
	Tianyi Tang\textsuperscript{\rm{2}}\footnotemark[1],
	Gaole He\textsuperscript{\rm{2}},
	Jinhao Jiang\textsuperscript{\rm{1}}, 
	\textbf{Xiaoxuan Hu}\textsuperscript{\rm{2}}, \\
	\textbf{Puzhao Xie}\textsuperscript{\rm{2}}, 
	\textbf{Zhipeng Chen}\textsuperscript{\rm{2}},
	\textbf{Zhuohao Yu}\textsuperscript{\rm{2}},
	\textbf{Wayne Xin Zhao}\textsuperscript{\rm{1,3}}\thanks{\ \ Corresponding author.}, 
	\textbf{Ji-Rong Wen}\textsuperscript{\rm{1,2,3}} \\
	
	\textsuperscript{1}Gaoling School of Artificial Intelligence, Renmin University of China \\
	\textsuperscript{2}School of Information, Renmin University of China \\
	\textsuperscript{3}Beijing Key Laboratory of Big Data Management and Analysis Methods \\
	\texttt{\{lijunyi,steven\_tang\}@ruc.edu.cn} \quad \texttt{batmanfly@gmail.com}
}

\date{}

\begin{document}
\maketitle
\begin{abstract}
In this paper, we release an open-source library, called \texttt{TextBox}, to provide a unified, modularized, and extensible text generation framework. TextBox aims to support a broad set of text generation tasks and models. In our library, we implement 21 text generation models on 9 benchmark datasets, covering the categories of VAE, GAN, and pretrained language models. Meanwhile, our library maintains sufficient modularity and extensibility by properly decomposing the model architecture, inference, and learning process into highly reusable modules, which allows users to easily incorporate new models into our framework. The above features make TextBox specially suitable for researchers and practitioners to quickly reproduce baseline models and develop new models. TextBox is implemented based on PyTorch, and released under Apache License 2.0 at the link \url{https://github.com/RUCAIBox/TextBox}.
\end{abstract}

\section{Introduction}

Text generation, which has emerged as an important branch of natural language processing (NLP), is often formally referred as natural language generation (NLG)~\cite{abs-2010-04389}. It aims to produce plausible and understandable text in human language from input data (\eg a sequence, keywords) or machine representation. Because of incredible performance of deep learning models, many classic text generation tasks have achieved rapid progress, such as machine translation~\cite{VaswaniSPUJGKP17}, dialogue systems~\cite{LiMRJGG16}, text summarization~\cite{SeeLM17}, text paraphrasing~\cite{MadnaniD10}, and more. 

To facilitate the development of text generation models, a few remarkable open-source libraries have been developed~\cite{BritzGLL17,KleinKDSR17,MillerFBBFLPW17,ZhuLZGZWY18,HuSTWYZHQWMLLZS19}. These frameworks are mainly designed for some or a small number of specific tasks, particularly machine translation and dialogue systems. They usually focus on a special kind of techniques for text generation such as generative adversarial networks (GAN), or have limitations in covering  commonly-used baseline implementations. Even for an experienced researcher, it is difficult and time-consuming to implement all compared baselines under a unified framework. Therefore, it is highly desirable to re-consider the implementation of text generation algorithms in a unified and modularized framework. 

In order to alleviate the above issues, we initiate a project to provide a unified framework for text generation algorithms. 
We implement an open-source text generation library, called \texttt{TextBox}, aiming to enhance the reproducibility of existing text generation models, standardize the implementation and evaluation protocol of text generation algorithms, and ease the development process of new algorithms. 
Our work is also useful to support several real-world applications in the field of text generation. We have extensively surveyed related text generation libraries and broadly fused their merits into TextBox. The key features and capabilities of our library are summarized in the following three aspects:

 \textbullet~Unified and modularized framework. TextBox is built upon PyTorch~\cite{PaszkeGMLBCKLGA19}, which is one of the most popular deep learning framework (especially in the research community). Moreover, it is designed to be highly modularized, by decoupling text generation models into a set of highly reusable modules, including data module, model module, evaluation module, and many common components and functionalities. In our library, it is convenient to compare different text generation algorithms with built-in evaluation protocols via simple yet flexible configurations, or develop new text generation models at a highly conceptual level by plugging in or swapping out modules.
   
 \textbullet~Comprehensive models, benchmark datasets and standardized evaluations. TextBox contains a wide range of text generation models, covering the categories of variational auto-encoder~(VAE), generative adversarial networks~(GAN), recurrent neural network~(RNN), Transformer-based models, and pretrained language models (PLM). We provide flexible supporting mechanisms via the configuration file or command line to run, compare and test these traditional and state-of-the-art algorithms. Based on these models, we implement two major text generation tasks, namely unconditional text generation tasks and conditional text generation tasks (\eg text summarization and  machine translation). To construct a reusable benchmark, we incorporate 9 widely-used datasets with regards to different text generation tasks for evaluation. Our library supports a series of frequently adopted evaluation protocols for testing and comparing text generation algorithms, such as perplexity, negative log-likelihood, BLEU, and ROUGE.
 
 \begin{figure*}[t]
	\centering
	\includegraphics[width=0.85\textwidth]{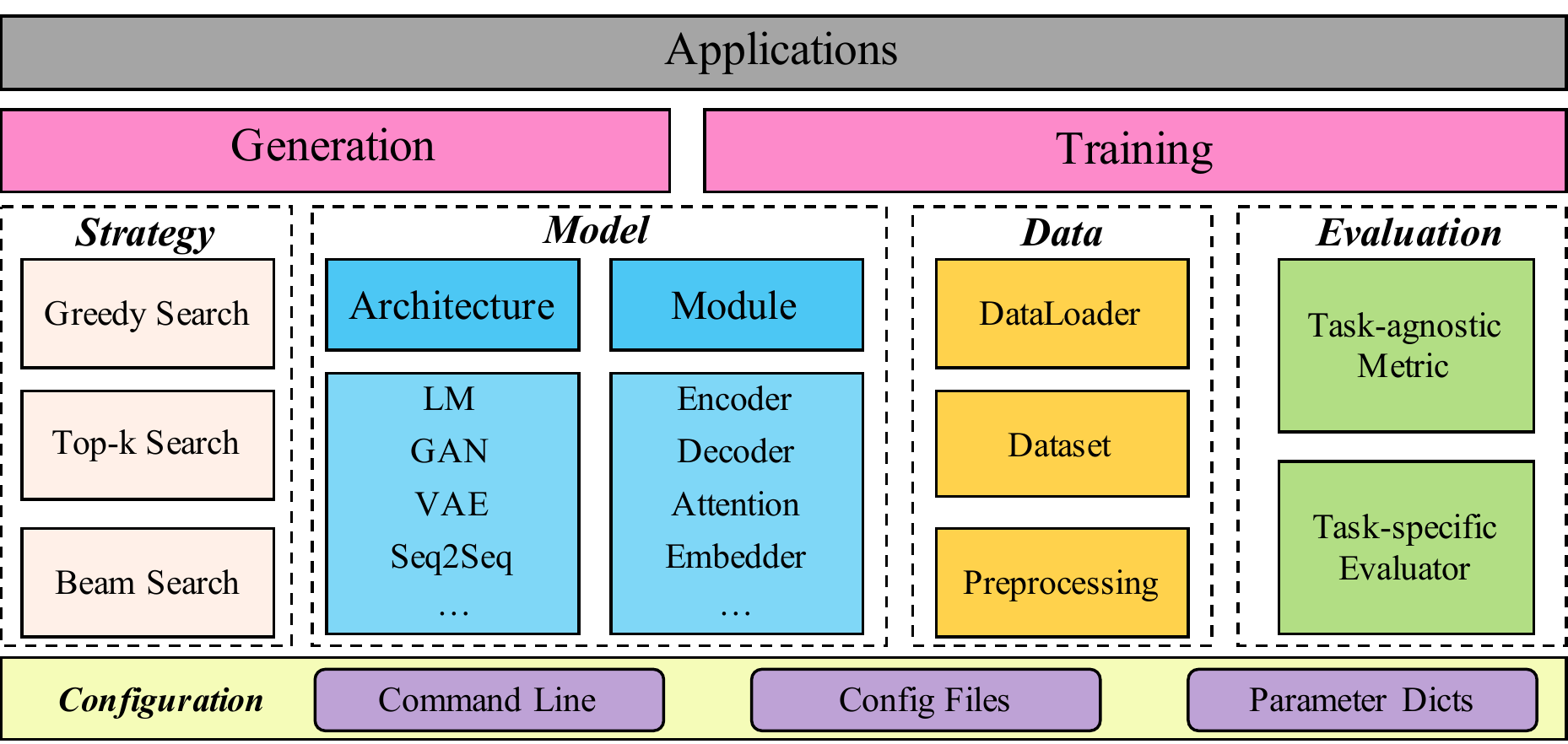}
	\caption{The illustration of the main functionalities and modules in our library TextBox.}
	\label{fig-framework}
\end{figure*}
 
 \textbullet~Extensible and flexible framework. TextBox provides convenient interfaces of various common functions or modules in text generation models, \eg RNN-based and Transformer-based encoders and decoders, pretrained language models, and attention mechanisms. Within our library, users are convenient to choose different API interfaces for building and evaluating their own models. Besides, the interfaces of our library are fully compatible with the PyTorch interface which allows seamless integration of user-customized modules and functions as needed.

\section{Architecture and Design}

Figure~\ref{fig-framework} presents the illustration of the main functionalities and modules in our library TextBox. The configuration module at the bottom helps users set up the experimental environment (\eg hyper-parameters and running details). Built upon the configuration module, the data, model, and evaluation modules form the core elements of our library. In the following, we describe the detailed structure of these three modules.

\subsection{Data Module}

A major design principle of our library is to support different text generation tasks. For this purpose, data module is the fundamental part to provide various data structures and functions adapting to different generation tasks.

For extensibility and reusability, our data module designs a unified data flow feeding input text into the models. The data flow can be described as: input text $\rightarrow$ \texttt{Dataset} $\rightarrow$ \texttt{DataLoader} $\rightarrow$ models. The class \texttt{Dataset} involves two special data structures, \ie single sequence and paired sequence, which are oriented to unconditional and conditional text generation tasks, respectively. The single sequence structure requires users to preprocess input text into one sequence per line in input files, while the paired sequence structure requires users to separate the source and target into two files with one sequence per line in each file. Specifically, for conditional text generation, TextBox supports several source formats corresponding to different tasks, \eg discrete attributes or tokens for attribute-to-text and keyword-to-text generation, a text sequence for machine translation or text summarization, and multiple text sequences for multi-turn dialogue systems. Furthermore, users can also provide additional information as inputs, \eg background text for agents in dialogues. The implementation of \texttt{Dataset} contains many common data preprocessing functionalities, such as converting text into lowercase, word tokenization, and building vocabulary. And the class \texttt{Dataloader} is based on the above two data structures, which is responsible for organizing the data stream.

In order to compare different generation models, we have collected 9 commonly-used benchmarks for text generation tasks, which makes it quite convenient for users to start with our library. 

\ignore{
\begin{table*}[t]
	\centering
	\begin{tabular}{c | c | c| r r r | c}
		\toprule[1pt]
		\multicolumn{2}{c|}{\textbf{Task}}&\textbf{Dataset}& \textbf{Train} & \textbf{Dev} & \textbf{Test}& \textbf{Structure}\\
		\midrule[0.7pt]
		\multicolumn{2}{c|}{\multirow{3}{*}{\tabincell{c}{Unconditional}}}&	COCO &	10,000&	10,000&	10,000& \multirow{3}{*}{\tabincell{c}{Single\\ Sequence}}\\
		\multicolumn{2}{c|}{}&EMNLP&	268,586&	10,000&	10,000& \\
		\multicolumn{2}{c|}{}&IMDB&	80,000&	10,000&	10,000& \\
		\midrule[0.7pt]
		\multirow{3}{*}{\tabincell{c}{Conditional}}&\multirow{2}{*}{\tabincell{c}{Translation}}&	IWSLT14 De$\rightarrow$EN&	153,348&	6,970&	6,750& \multirow{3}{*}{\tabincell{c}{Paired \\ Sequence}}\\
		&	& WMT14 En$\rightarrow$De&	3,621,184&	36,652&	2,931& \\
		\cline{2-6}
		&\multirow{1}{*}{\tabincell{c}{Summarization}}&	GigaWord&	3,803,957&	189,651&	1,936&\\
		\bottomrule[1pt]
	\end{tabular}%
	\caption{Benchmarks used in our library TextBox.}
	\label{tab-data}%
\end{table*}}

\subsection{Model Module}

To support a variety of models, we set up the model module by decoupling the algorithm implementation from other components and abstracting a set of widely-used modules, \eg \texttt{encoder}, \texttt{decoder}. These modules can be flexibly combined following the required interface and then connected with data and evaluation modules. Based on this abstract design, it is convenient to switch between different text generation tasks, and change from one modeling paradigm to another by simply plugging in or swapping out modules. 

In addition to modularized design, our library also includes a large number of text generation baseline models for  reproducibility. At the first released version, we have implemented 21 baseline models within three main categories of text generation models, namely VAE-based, GAN-based, and pretrained language models, corresponding to different generation architectures and tasks. For example, GAN-based models consist of \texttt{generator} and \texttt{discriminator}, and VAE-based models contain \texttt{encoder} and \texttt{decoder}. We summarize all the implemented models in Table~\ref{tab-model}. For all the implemented models, we test their performance for unconditional and conditional generation tasks on corresponding benchmarks, and invite a code reviewer to examine the correctness of the implementation. Overall, the extensible and comprehensive model modules can be beneficial for fast exploration of new algorithms for a specific task, and convenient comparison between different models.

In specific, for each model, we utilize two interface functions, \ie \texttt{forward} and \texttt{generate}, for training and testing, respectively. These functions are general to various text generation algorithms, so that we can implement various algorithms in a highly unified way. Such a design also enables quick development of new models.

In order to improve the quality of generation results, we also implement a series of generation strategies when generating text, such as greedy search, top-$k$ search and beam search. Users are allowed to switch between different generation strategies leading to better performance through setting a hyper-parameter, \ie \texttt{decoding\_strategy}. Besides, we add the functions of model saving and loading to store and reuse the learned models, respectively. In the training process, one can print and monitor the change of the loss value and apply training tricks such as warm-up and early-stopping. These tiny tricks largely improve the usage experiences with our library.

\begin{table}[t]
	\centering
	\small
	\begin{tabular}{c| c | c}
		\toprule[1pt]
		\textbf{Category}& \textbf{Models} & \textbf{Reference}\\
		\midrule[0.7pt]
		\multirow{4}{*}{\tabincell{c}{VAE}}&	LSTM-VAE&	\cite{Bowman-SIGNLL-2016}\\
		&	CNN-VAE&	\cite{Yang-ICML-2017}\\
		&	Hybrid-VAE&	\cite{Semeniuta-EMNLP-2017}\\
		&	CVAE& \cite{LiSZCSZY18} \\
		\midrule[0.7pt]
		\multirow{6}{*}{\tabincell{c}{GAN}}&	SeqGAN&	\cite{Yu-AAAI-2017}\\
		&	TextGAN&	\cite{Zhang-ICML-2017}\\
		&	RankGAN&	\cite{Lin-NIPS-2017}\\
		&	MaliGAN&	\cite{Che-MaliGAN-2017}\\
		&	LeakGAN&	\cite{Guo-AAAI-2018}\\
		&	MaskGAN&	\cite{Fedus-ICLR-2018}\\
		\midrule[0.7pt]
		\multirow{6}{*}{\tabincell{c}{Pretrained\\Language\\Model}}&GPT-2 	&	\cite{radford2019language}\\
		&XLNet 	&	\cite{Yang-NIPS-2019}\\
		&BERT2BERT 	&	\cite{Rothe-TACL-2020}\\
		&BART 	&	\cite{Lewis-ACL-2020}\\
		&ProphetNet & \cite{QiYGLDCZ020} \\
		&T5 & \cite{RaffelSRLNMZLL20} \\
		\midrule[0.7pt]
		\multirow{5}{*}{\tabincell{c}{Seq2Seq}}&RNN 	&	\cite{Sutskever-NIPS-2014}\\
		& Transformer & \cite{VaswaniSPUJGKP17} \\
		& Context2Seq & \cite{TangYCZM16} \\
		& Attr2Seq & \cite{ZhouLWDHX17} \\
		& HRED & \cite{SerbanSBCP16} \\
		\bottomrule[1pt]
	\end{tabular}%
	\caption{Implemented models in our library TextBox.}
	\label{tab-model}%
	\vspace{-0.1cm}
\end{table}

\subsection{Evaluation Module}

It is important that different models should be compared under the unified evaluate protocols, which is useful to standardize the evaluation of text generation. To achieve this goal, we set up the evaluation module to implement commonly-used evaluation protocols for text generation models.

Our library supports both logit-based and word-based evaluation metrics. The logit-based metrics (for unconditional text generation task) include negative log-likelihood (NLL)~\cite{Huszar15} and perplexity (PPL)~\cite{BrownPPLM92}, measuring how well the probability distribution or a probability model predicts a sample compared with the ground-truth. The word-based metrics (for both unconditional and conditional text generation tasks) include the most widely-used generation metrics. For example, BLEU-$n$~\cite{PapineniRWZ02} and ROUGE-$n$~\cite{lin2004rouge} measure the ratios of the overlapping $n$-grams between the generated and real samples, and Distinct-$n$~\cite{LiGBGD16} mesures the degree of diversity of generated text by calculating the number of distinct unigrams and bigrams in generated text. Besides, to evaluate the diversity of generated samples, we also take into account the Self-BLEU~\cite{ZhuLZGZWY18} metric in text generation. In summary, users can choose different evaluation protocols towards a specific generation task by setting the hyper-parameter, \ie \texttt{metrics}.

In practice, as the model may generate many text pieces, evaluation efficiency is an important concern. Hence, we integrate efficient computing package, \texttt{fastBLEU}~\cite{alihosseini-etal-2019-jointly}, to compute evaluation scores. Compared with other package, \texttt{fastBLEU} adopts the multi-threaded C++ implementation.

\section{System Usage}

In this section, we show a detailed guideline to use our system library. Users can run the existing models or add their own models as needed.

\subsection{Running Existing Models}

To run an existing model within TextBox, users only need to specify the dataset and model by setting hyper-parameters, \ie \texttt{dataset} and \texttt{model}. And then experiments can be run with a simple command-line interface:\\

\texttt{python run\_textbox.py \textbackslash}

\texttt{~~-{}-model=GPT2 -{}-dataset=COCO }\\

The above case shows an example that runs GPT-2~\cite{radford2019language} model on COCO dataset~\cite{lin2015microsoft}. In our system library, the generation task, such as \texttt{translation}, and \texttt{summarization}, is determined once users specify the dataset, thus the task is not necessary to be explicitly specified in hyper-parameters. To facilitate the modification of hyper-parameters, we provides two kinds of YAML configuration files, \ie dataset configuration and model configuration, which allow running many experiments without modifying source code. It also supports users to include hyper-parameters in the command line, which is useful for some specifically defined parameters. TextBox is designed to be run on different hardware devices. By default, CUDA devices will be used if users set the hyper-parameter \texttt{use\_gpu} as \texttt{True}, or otherwise CPU will be used. Users can determine the ID of used CUDA devices by setting hyper-parameter \texttt{gpu\_id}. We also support distributed model training in multiple GPUs by setting the hyper-parameter \texttt{DDP} as \texttt{True}.

Based on the configuration, we provide the auxiliary function to split the dataset into train, validation and test sets according to the provided hyper-parameter \texttt{split\_ratio}, or load the pre-splitted dataset. Moreover, TextBox also allows users to load and re-train the saved model for speeding up reproduction, rather than training from scratch.

\begin{figure}[t]
	\centering
	\includegraphics[width=0.43\textwidth]{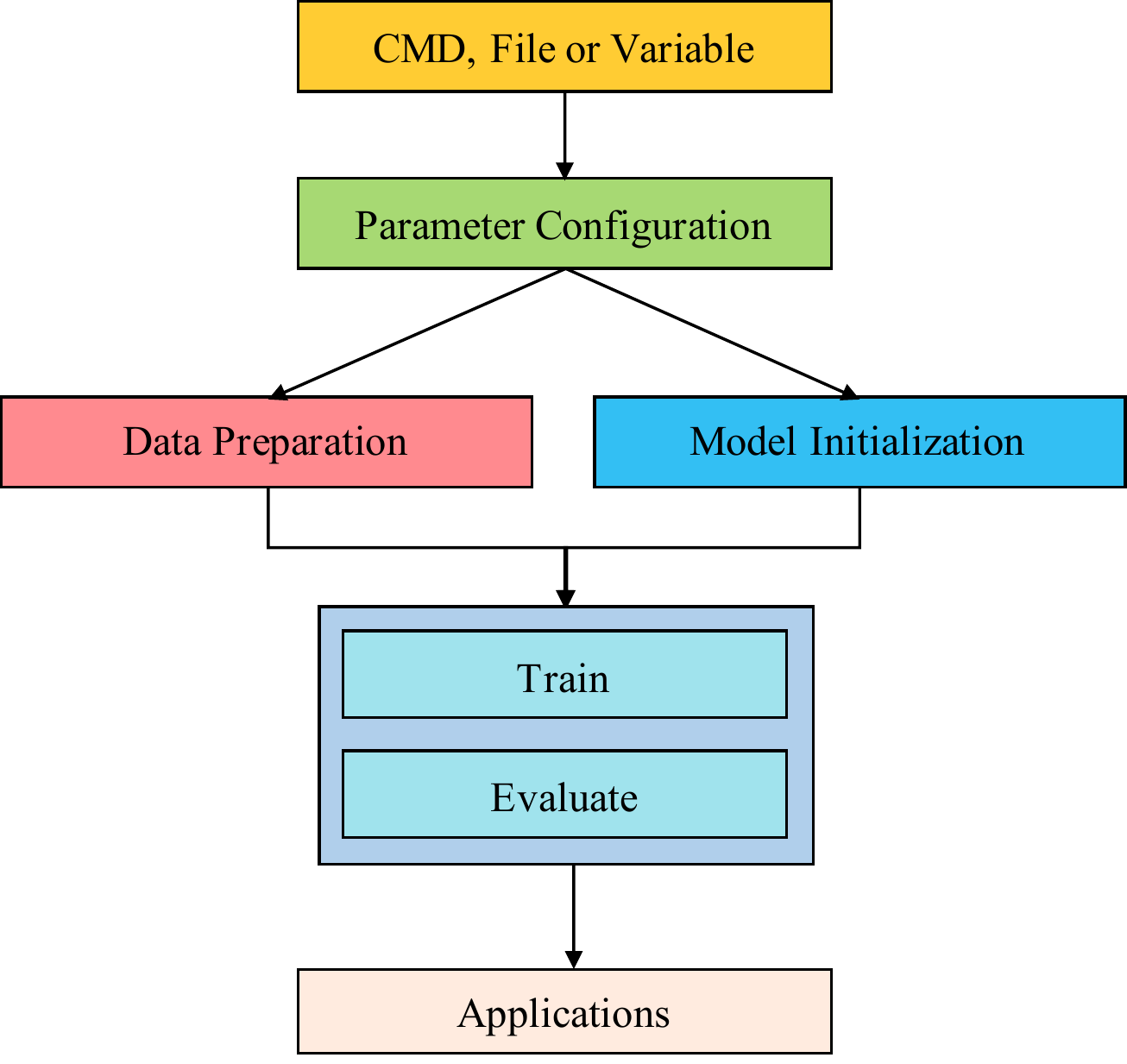}
	\caption{An illustractive usage flow of our library.}
	\label{fig-case}
\end{figure}

Figure~\ref{fig-case} presents a general usage flow when running a model in our library. The running procedure relies on some experimental configuration, obtained from the files, command line or parameter dictionaries. The dataset and model are prepared and initialized according to the configured settings, and the execution module is responsible for training and evaluating models.

\subsection{Implementing a New Model}
With the unified \texttt{Data} and \texttt{Evaluation} modules, one needs to implement a specific \texttt{Model} class and three mandatory functions as follows:

\textbullet~\texttt{\_\_init\_\_()} function. In this function, the user performs parameters initialization, global variable definition and so on. It is worth noting that, the imported new model should be a sub-class of the abstract model class defined in our library. One can reuse the modules (\eg Transformer) and layers (\eg Highway net) already existing in our library for convenience. A configuration file is preferable to conduct further flexible adjustment. 

\textbullet~\texttt{forward()} function. This function calculates the training loss to be optimized and validation loss to avoid overfitting. Based on the returned training loss, our library will automatically invoke different optimization methods to learn the parameters according to pre-defined configuration.

\textbullet~\texttt{generate()} function. This function is employed to generate output text based on input text or free text. Our library also provides several generation strategies, such as beam search and top-$k$ search, for users to improve generation results.

In order to implement user-customized modules, one can reuse functions and classes inherited from our basic modules, or override original functions and add new functions.

\section{Performance Evaluation}

To evaluate the models in TextBox, we conduct extensive experiments to compare their performance on unconditional and conditional generation tasks. 

\begin{table*}[t]
	\renewcommand\arraystretch{1.1}
	\small
	\begin{center}
		\begin{tabular}{c|c|l | c c | c c c c}
			\toprule[1pt]
			\textbf{Tasks} & \textbf{Datasets}& \textbf{Models}& \textbf{Distinct-1}&  \textbf{Distinct-2} & \textbf{BLEU-1}& \textbf{BLEU-2}& \textbf{BLEU-3} & \textbf{BLEU-4} \\
			\midrule[0.7pt]
			\multirow{5}{*}{\tabincell{c}{Unconditional\\Generation}}&\multirow{5}{*}{COCO}& LSTM-VAE & -& -& 63.97 & 46.56 & 18.53 & 5.97 \\
			&& SeqGAN  & -& -& 99.76 & 82.32 & 51.26 & 25.18  \\
			&& RankGAN & -& -& 99.76 & 82.92 & 52.46 & 26.40 \\
			&& MailGAN & -& -& 99.71 & 81.95 & 50.86 & 24.87 \\
			&& GPT-2  & -& -& 88.15 & 78.13 & 55.81 & 31.88 \\
			\midrule[0.7pt]
			\multirow{2}{*}{\tabincell{c}{Attribute-to-Text\\Generation}}&\multirow{2}{*}{AMAZON}& Context2Seq  & 0.07 & 0.39 & 17.21 & 2.80 & 0.83 & 0.43  \\
			&& Attr2Seq & 0.14 & 2.81 & 17.14 & 2.81 & 0.87 & 0.48\\
			\midrule[0.7pt]
			\multirow{3}{*}{\tabincell{c}{Dialogue\\Systems}}&\multirow{3}{*}{\tabincell{c}{Personal\\Chat}}& RNN+Attn  & 0.24 & 0.72 & 17.51 & 4.65 & 2.11 & 1.47 \\
			&& Transformer & 0.38 & 2.28 & 17.29 & 4.85 & 2.32 & 1.65 \\
			&& HRED & 0.22 & 0.63 & 17.29 & 4.72 & 2.20 & 1.60 \\
			\bottomrule[1pt]
		\end{tabular}
		\caption{Performance comparisons of different methods for three tasks, \ie unconditional generation, attribute-to-text generation, and dialogue systems. Distinct-$n$ is not applicable to the unconditional generation task.}
		\label{tab:unconditional_results}
	\end{center}
\end{table*}

\subsection{Unconditional Text Generation}

Following previous work, we adopt COCO~\cite{lin2015microsoft}, EMNLP2017 WMT News~\cite{chatterjee-EtAl:2017:WMT1} and IMDB Movie Reviews~\cite{maas-EtAl:2011:ACL-HLT2011} datasets for comparing the performance of five traditional and state-of-the-art models, \ie LSTM-VAE, SeqGAN, RankGAN, MaliGAN, and GPT-2, in the unconditional text generation task. 

In our experiments, we run models with the parameter configurations described in their original papers. Note that the BLEU-$n$ metric employs the one-hot weights (\eg $(0,0,0,1)$ for BLEU-4) instead of average weights, since we consider that one-hot weights can reflect the overlapping $n$-grams more realistically. 

These results on COCO datasets are shown in Table~\ref{tab:unconditional_results}, and other results on EMNLP2017 and IMDB datasets can be found in our GitHub page. We can see from Table~\ref{tab:unconditional_results}, these models implemented in our library have the comparable performance compared with the results reported in the original papers. Moreover, the pretrained language model, \ie GPT-2, achieves consistent and remarkable performance, which is line with our expectations.

\subsection{Conditional Text Generation}

In this section, we apply various models on four conditional text generation tasks, \ie attribute-to-text generation, dialogue systems, machine translation, and text summarization. The task of attribute-to-text generation is to generate text given several discrete attributes, such as user, item, and rating. We use the popular context-to-sequence (Context2Seq) and attribute-to-sequence (Attr2Seq) as base models, which utilize the multi-layer perceptron (MLP) and RNN as the encoder and decoder, respectively. Besides, dialogue systems aim to generate response given a conversation history. We consider two typical models, \ie Attention-based RNN and Transformer, and one popular hierarchical recurrent encoder-decoder model (HRED) as base models. In RNN and Transformer, the multi-sequence conversation history is concatenated as one sequence feeding into the encoder, while in HERD the hierarchical structure of the conversation history is kept and modeled with a hierarchical encoder. Their results are shown in Table~\ref{tab:unconditional_results}.

\begin{table}[tbp]
	\centering
	\small
	\begin{tabular}{c|l|r r r}
		\toprule[1pt]
		\textbf{Model}&	\textbf{Strategy}& \textbf{BLEU2} & \textbf{BLEU3} & \textbf{BLEU4}\\
		\midrule[0.7pt]
		\multirow{3}{*}{\tabincell{c}{RNN+Attn}}
		&\textbf{Top-$k$} & 26.68 & 16.95 & 10.85 \\
		&\textbf{Greedy} & 33.74 & 23.03 & 15.79 \\
		&\textbf{Beam} & 35.68 & 24.94 & 17.42 \\
		\midrule[0.7pt]
		\multirow{3}{*}{\tabincell{c}{Transformer}}
		&\textbf{Top-$k$} & 30.96 & 20.83 & 14.16 \\
		&\textbf{Greedy} & 35.48 & 24.76 & 17.41 \\
		&\textbf{Beam} & 36.88 & 26.10 & 18.54 \\
		\bottomrule[1pt]
	\end{tabular}%
	\caption{Performance comparison of different generation models with three  strategies for machine translation from German to English.}
	\label{tab:translation_results}%
\end{table}

 To showcase how our TextBox can support diverse techniques on several tasks with different decoding strategies, we compare the attention-based RNN model, Transformer, and four state-of-the-art pretrained language models, \ie BART, BERT2BERT, ProphetNet, and T5, for both machine translation and text summarization tasks. In Table~\ref{tab:translation_results}, we adopt the IWSLT2014 German-to-English~\cite{cettolo2014report} translation dataset and utilize three generation strategies, \ie top-$k$, greedy, and beam search. The greedy strategy considers the most probable token at each generation step, the top-$k$ search strategy means sorting by probability and zero-ing out the probabilities for anything below the $k$-th token, and beam search~\cite{VijayakumarCSSL18} strategy selects the top scoring $B$ candidates from the set of all possible one token extensions of its beams, where $B$ is the beam size ($B=5$ in our experiments). From Table~\ref{tab:translation_results} we observe that the beam search strategy brings more improvement than the others. For text summarization, we compare RNN and Transformer with four pretrained models as shown in Table~\ref{tab:summarization_results}. These models are trained or fine-tuned in GigaWord~\cite{graff2003english} dataset. As observed in Table~\ref{tab:summarization_results}, pretrained models outperform the RNN model and Transformer by a clear margin. 
 
 The results of all implemented models in other tasks and datasets can be acquired from our GitHub page.

\section{Related Work}

Several toolkits have been released focusing on one or a few specific text generation tasks or techniques. For example, Tensor2Tensor~\cite{vaswani-etal-2018-tensor2tensor}, MarianNMT~\cite{mariannmt} and OpenNMT~\cite{klein-etal-2017-opennmt} are designed for machine translation task, while ParlAI~\cite{miller-etal-2017-parlai} and Plato~\cite{papangelis2020plato} specialized for dialog research in this field. There are two text generation libraries closely related to our library, including Texygen~\cite{ZhuLZGZWY18} and Texar~\cite{HuSTWYZHQWMLLZS19} focusing on GAN technique and high modularization, respectively. TextBox has drawn inspirations from these toolkits when designing relevant functions.

Compared with them, TextBox covers more text generation tasks and models, which is useful for reproducibility. Besides, we implement standardized evaluation to compare different models. Also, our library provides various common modules for convenience. It has a proper focus on text generation field, and provide a comprehensive set of modules and functionalities.
\section{Conclusion}

\begin{table}[tbp]
	\centering
	\small
	\begin{tabular}{l|r r r}
		\toprule[1pt]
		\textbf{Model}&	\textbf{ROUGE-1}& \textbf{ROUGE-2} & \textbf{ROUGE-L} \\
		\midrule[0.7pt]
		RNN+Attn & 36.32 & 17.63 & 38.36\\
		Transformer & 36.21 & 17.64 & 38.10\\
		\midrule[0.7pt]
		BART & 39.34 & 20.07 & 41.25 \\
		BERT2BERT & 38.16 & 18.89 & 40.06 \\
		ProphetNet & 38.49 & 18.41 & 39.84  \\
		T5 & 38.83 & 19.68 & 40.76 \\
		\bottomrule[1pt]
	\end{tabular}%
	\caption{Performance comparison of different generation models for text summarization. Specifically, we adopt the base version of BART, BERT2BERT, T5 and the large version of ProphetNet.}
	\label{tab:summarization_results}%
\end{table}

This paper presented a unified, modularized, and extensible text generation library, called \texttt{TextBox}. So far, we have implemented 21 text generation models, including VAE-based, GAN-based, RNN-based  Transformer-based models and pretrained language models, and 9 benchmark datasets for unconditional and conditional text generation tasks. Moreover, Our library is modularized to easily plug in or swap out components, and extensible to support seamless incorporation of other external modules. In the future, features and functionalities will continue be added to our library, including more models and datasets, diverse inputs such as graph and table, and distributed training in multiple machines. We invite researchers and practitioners to join and enrich TextBox, and help push forward the text generation research.

\section{Broader Impacts}

Text generation has a wide range of beneficial applications for society, including code auto-completion, game narrative generation, and answering questions. But it also has potentially harmful applications. For example, GPT-3 improves the quality of generated text over smaller models and increases the difficulty of distinguishing synthetic text from human-written text, such as fake news and reviews. 

Here we focus on two potential issues: the potential for deliberate misuse of generation models and the issue of bias. Malicious uses of generation models can be somewhat difficult to anticipate because they often involve repurposing models in a very different environment or for a different purpose than researchers intended. To mitigate this, we can think in terms of traditional security risk assessment frameworks such as identifying threats. Biases present in training text may lead models to generate stereotyped or prejudiced content. This is concerning, since model bias could harm people in the relevant groups in different ways.  In order to prevent bias, there is a need for building a common vocabulary tying together the normative, technical and empirical challenges of bias mitigation for generation models. We expect this to be an area of continuous research for us.

\bibliography{textbox_bib}

\begin{thebibliography}{48}
\expandafter\ifx\csname natexlab\endcsname\relax\def\natexlab#1{#1}\fi

\bibitem[{Alihosseini et~al.(2019)Alihosseini, Montahaei, and
  Soleymani~Baghshah}]{alihosseini-etal-2019-jointly}
Danial Alihosseini, Ehsan Montahaei, and Mahdieh Soleymani~Baghshah. 2019.
\newblock \href {https://doi.org/10.18653/v1/W19-2311} {Jointly measuring
  diversity and quality in text generation models}.
\newblock In \emph{Proceedings of the Workshop on Methods for Optimizing and
  Evaluating Neural Language Generation}, pages 90--98, Minneapolis, Minnesota.
  Association for Computational Linguistics.

\bibitem[{Bowman et~al.(2016)Bowman, Vilnis, Vinyals, Dai, J{\'{o}}zefowicz,
  and Bengio}]{Bowman-SIGNLL-2016}
Samuel~R. Bowman, Luke Vilnis, Oriol Vinyals, Andrew~M. Dai, Rafal
  J{\'{o}}zefowicz, and Samy Bengio. 2016.
\newblock Generating sentences from a continuous space.
\newblock In \emph{Proceedings of the 20th {SIGNLL} Conference on Computational
  Natural Language Learning, CoNLL 2016, Berlin, Germany, August 11-12, 2016},
  pages 10--21.

\bibitem[{Britz et~al.(2017)Britz, Goldie, Luong, and Le}]{BritzGLL17}
Denny Britz, Anna Goldie, Minh-Thang Luong, and Quoc Le. 2017.
\newblock Massive exploration of neural machine translation architectures.
\newblock In \emph{Proceedings of the 2017 Conference on Empirical Methods in
  Natural Language Processing}, pages 1442--1451, Copenhagen, Denmark.
  Association for Computational Linguistics.

\bibitem[{Brown et~al.(1992)Brown, Pietra, Pietra, Lai, and
  Mercer}]{BrownPPLM92}
Peter~F. Brown, Stephen~Della Pietra, Vincent J.~Della Pietra, Jennifer~C. Lai,
  and Robert~L. Mercer. 1992.
\newblock An estimate of an upper bound for the entropy of english.
\newblock \emph{Comput. Linguistics}, 18(1):31--40.

\bibitem[{Cettolo et~al.(2014)Cettolo, Niehues, St{\"u}ker, Bentivogli, and
  Federico}]{cettolo2014report}
Mauro Cettolo, Jan Niehues, Sebastian St{\"u}ker, Luisa Bentivogli, and
  Marcello Federico. 2014.
\newblock Report on the 11th iwslt evaluation campaign, iwslt 2014.
\newblock In \emph{Proceedings of the International Workshop on Spoken Language
  Translation, Hanoi, Vietnam}, volume~57.

\bibitem[{Chatterjee et~al.(2017)Chatterjee, Negri, Turchi, Federico, Specia,
  and Blain}]{chatterjee-EtAl:2017:WMT1}
Rajen Chatterjee, Matteo Negri, Marco Turchi, Marcello Federico, Lucia Specia,
  and Fr\'{e}d\'{e}ric Blain. 2017.
\newblock \href {http://www.aclweb.org/anthology/W17-4716} {Guiding neural
  machine translation decoding with external knowledge}.
\newblock In \emph{Proceedings of the Second Conference on Machine Translation,
  Volume 1: Research Papers}, pages 157--168, Copenhagen, Denmark. Association
  for Computational Linguistics.

\bibitem[{Che et~al.(2017)Che, Li, Zhang, Hjelm, Li, Song, and
  Bengio}]{Che-MaliGAN-2017}
Tong Che, Yanran Li, Ruixiang Zhang, R.~Devon Hjelm, Wenjie Li, Yangqiu Song,
  and Yoshua Bengio. 2017.
\newblock Maximum-likelihood augmented discrete generative adversarial
  networks.
\newblock \emph{CoRR}, abs/1702.07983.

\bibitem[{Dong et~al.(2017)Dong, Huang, Wei, Lapata, Zhou, and
  Xu}]{ZhouLWDHX17}
Li~Dong, Shaohan Huang, Furu Wei, Mirella Lapata, Ming Zhou, and Ke~Xu. 2017.
\newblock Learning to generate product reviews from attributes.
\newblock In \emph{Proceedings of the 15th Conference of the European Chapter
  of the Association for Computational Linguistics, {EACL} 2017, Valencia,
  Spain, April 3-7, 2017, Volume 1: Long Papers}, pages 623--632. Association
  for Computational Linguistics.

\bibitem[{Fedus et~al.(2018)Fedus, Goodfellow, and Dai}]{Fedus-ICLR-2018}
William Fedus, Ian~J. Goodfellow, and Andrew~M. Dai. 2018.
\newblock Maskgan: Better text generation via filling in the
  {\_}{\_}{\_}{\_}{\_}{\_}{\_}.
\newblock In \emph{6th International Conference on Learning Representations,
  {ICLR} 2018, Vancouver, BC, Canada, April 30 - May 3, 2018, Conference Track
  Proceedings}.

\bibitem[{Graff et~al.(2003)Graff, Kong, Chen, and Maeda}]{graff2003english}
David Graff, Junbo Kong, Ke~Chen, and Kazuaki Maeda. 2003.
\newblock English gigaword.
\newblock \emph{Linguistic Data Consortium, Philadelphia}, 4(1):34.

\bibitem[{Guo et~al.(2018)Guo, Lu, Cai, Zhang, Yu, and Wang}]{Guo-AAAI-2018}
Jiaxian Guo, Sidi Lu, Han Cai, Weinan Zhang, Yong Yu, and Jun Wang. 2018.
\newblock Long text generation via adversarial training with leaked
  information.
\newblock In \emph{Proceedings of the Thirty-Second {AAAI} Conference on
  Artificial Intelligence, (AAAI-18), the 30th innovative Applications of
  Artificial Intelligence (IAAI-18), and the 8th {AAAI} Symposium on
  Educational Advances in Artificial Intelligence (EAAI-18), New Orleans,
  Louisiana, USA, February 2-7, 2018}, pages 5141--5148.

\bibitem[{Hu et~al.(2019)Hu, Shi, Tan, Wang, Yang, Zhao, He, Qin, Wang, Ma,
  Liu, Liang, Zhu, Sachan, and Xing}]{HuSTWYZHQWMLLZS19}
Zhiting Hu, Haoran Shi, Bowen Tan, Wentao Wang, Zichao Yang, Tiancheng Zhao,
  Junxian He, Lianhui Qin, Di~Wang, Xuezhe Ma, Zhengzhong Liu, Xiaodan Liang,
  Wanrong Zhu, Devendra~Singh Sachan, and Eric~P. Xing. 2019.
\newblock Texar: {A} modularized, versatile, and extensible toolkit for text
  generation.
\newblock In \emph{Proceedings of the 57th Conference of the Association for
  Computational Linguistics, {ACL} 2019, Florence, Italy, July 28 - August 2,
  2019, Volume 3: System Demonstrations}, pages 159--164. Association for
  Computational Linguistics.

\bibitem[{Huszar(2015)}]{Huszar15}
Ferenc Huszar. 2015.
\newblock \href {http://arxiv.org/abs/1511.05101} {How (not) to train your
  generative model: Scheduled sampling, likelihood, adversary?}
\newblock \emph{CoRR}, abs/1511.05101.

\bibitem[{Junczys-Dowmunt et~al.(2018)Junczys-Dowmunt, Grundkiewicz, Dwojak,
  Hoang, Heafield, Neckermann, Seide, Germann, Fikri~Aji, Bogoychev, Martins,
  and Birch}]{mariannmt}
Marcin Junczys-Dowmunt, Roman Grundkiewicz, Tomasz Dwojak, Hieu Hoang, Kenneth
  Heafield, Tom Neckermann, Frank Seide, Ulrich Germann, Alham Fikri~Aji,
  Nikolay Bogoychev, Andr\'{e} F.~T. Martins, and Alexandra Birch. 2018.
\newblock Marian: Fast neural machine translation in {C++}.
\newblock In \emph{Proceedings of ACL 2018, System Demonstrations}, pages
  116--121, Melbourne, Australia. Association for Computational Linguistics.

\bibitem[{Klein et~al.(2017{\natexlab{a}})Klein, Kim, Deng, Senellart, and
  Rush}]{klein-etal-2017-opennmt}
Guillaume Klein, Yoon Kim, Yuntian Deng, Jean Senellart, and Alexander Rush.
  2017{\natexlab{a}}.
\newblock {O}pen{NMT}: Open-source toolkit for neural machine translation.
\newblock In \emph{Proceedings of {ACL} 2017, System Demonstrations}, pages
  67--72, Vancouver, Canada. Association for Computational Linguistics.

\bibitem[{Klein et~al.(2017{\natexlab{b}})Klein, Kim, Deng, Senellart, and
  Rush}]{KleinKDSR17}
Guillaume Klein, Yoon Kim, Yuntian Deng, Jean Senellart, and Alexander~M. Rush.
  2017{\natexlab{b}}.
\newblock Opennmt: Open-source toolkit for neural machine translation.
\newblock In \emph{Proceedings of the 55th Annual Meeting of the Association
  for Computational Linguistics, {ACL} 2017, Vancouver, Canada, July 30 -
  August 4, System Demonstrations}, pages 67--72. Association for Computational
  Linguistics.

\bibitem[{Lewis et~al.(2020)Lewis, Liu, Goyal, Ghazvininejad, Mohamed, Levy,
  Stoyanov, and Zettlemoyer}]{Lewis-ACL-2020}
Mike Lewis, Yinhan Liu, Naman Goyal, Marjan Ghazvininejad, Abdelrahman Mohamed,
  Omer Levy, Veselin Stoyanov, and Luke Zettlemoyer. 2020.
\newblock {BART:} denoising sequence-to-sequence pre-training for natural
  language generation, translation, and comprehension.
\newblock In \emph{Proceedings of the 58th Annual Meeting of the Association
  for Computational Linguistics, {ACL} 2020, Online, July 5-10, 2020}, pages
  7871--7880.

\bibitem[{Li et~al.(2016{\natexlab{a}})Li, Galley, Brockett, Gao, and
  Dolan}]{LiGBGD16}
Jiwei Li, Michel Galley, Chris Brockett, Jianfeng Gao, and Bill Dolan.
  2016{\natexlab{a}}.
\newblock A diversity-promoting objective function for neural conversation
  models.
\newblock In \emph{{NAACL} {HLT} 2016, The 2016 Conference of the North
  American Chapter of the Association for Computational Linguistics: Human
  Language Technologies, San Diego California, USA, June 12-17, 2016}, pages
  110--119. The Association for Computational Linguistics.

\bibitem[{Li et~al.(2016{\natexlab{b}})Li, Monroe, Ritter, Jurafsky, Galley,
  and Gao}]{LiMRJGG16}
Jiwei Li, Will Monroe, Alan Ritter, Dan Jurafsky, Michel Galley, and Jianfeng
  Gao. 2016{\natexlab{b}}.
\newblock Deep reinforcement learning for dialogue generation.
\newblock In \emph{Proceedings of the 2016 Conference on Empirical Methods in
  Natural Language Processing, {EMNLP} 2016, Austin, Texas, USA, November 1-4,
  2016}, pages 1192--1202. The Association for Computational Linguistics.

\bibitem[{Li et~al.(2018)Li, Song, Zhang, Chen, Shi, Zhao, and
  Yan}]{LiSZCSZY18}
Juntao Li, Yan Song, Haisong Zhang, Dongmin Chen, Shuming Shi, Dongyan Zhao,
  and Rui Yan. 2018.
\newblock Generating classical chinese poems via conditional variational
  autoencoder and adversarial training.
\newblock In \emph{Proceedings of the 2018 Conference on Empirical Methods in
  Natural Language Processing, Brussels, Belgium, October 31 - November 4,
  2018}, pages 3890--3900. Association for Computational Linguistics.

\bibitem[{Lin(2004)}]{lin2004rouge}
Chin-Yew Lin. 2004.
\newblock Rouge: A package for automatic evaluation of summaries.
\newblock In \emph{Text summarization branches out}, pages 74--81.

\bibitem[{Lin et~al.(2017)Lin, Li, He, Sun, and Zhang}]{Lin-NIPS-2017}
Kevin Lin, Dianqi Li, Xiaodong He, Ming{-}Ting Sun, and Zhengyou Zhang. 2017.
\newblock Adversarial ranking for language generation.
\newblock In \emph{Advances in Neural Information Processing Systems 30: Annual
  Conference on Neural Information Processing Systems 2017, December 4-9, 2017,
  Long Beach, CA, {USA}}, pages 3155--3165.

\bibitem[{Lin et~al.(2015)Lin, Maire, Belongie, Bourdev, Girshick, Hays,
  Perona, Ramanan, Zitnick, and Dollár}]{lin2015microsoft}
Tsung-Yi Lin, Michael Maire, Serge Belongie, Lubomir Bourdev, Ross Girshick,
  James Hays, Pietro Perona, Deva Ramanan, C.~Lawrence Zitnick, and Piotr
  Dollár. 2015.
\newblock \href {http://arxiv.org/abs/1405.0312} {Microsoft coco: Common
  objects in context}.

\bibitem[{Maas et~al.(2011)Maas, Daly, Pham, Huang, Ng, and
  Potts}]{maas-EtAl:2011:ACL-HLT2011}
Andrew~L. Maas, Raymond~E. Daly, Peter~T. Pham, Dan Huang, Andrew~Y. Ng, and
  Christopher Potts. 2011.
\newblock \href {http://www.aclweb.org/anthology/P11-1015} {Learning word
  vectors for sentiment analysis}.
\newblock In \emph{Proceedings of the 49th Annual Meeting of the Association
  for Computational Linguistics: Human Language Technologies}, pages 142--150,
  Portland, Oregon, USA. Association for Computational Linguistics.

\bibitem[{Madnani and Dorr(2010)}]{MadnaniD10}
Nitin Madnani and Bonnie~J. Dorr. 2010.
\newblock Generating phrasal and sentential paraphrases: {A} survey of
  data-driven methods.
\newblock \emph{Comput. Linguistics}, 36(3):341--387.

\bibitem[{Miller et~al.(2017{\natexlab{a}})Miller, Feng, Batra, Bordes, Fisch,
  Lu, Parikh, and Weston}]{miller-etal-2017-parlai}
Alexander Miller, Will Feng, Dhruv Batra, Antoine Bordes, Adam Fisch, Jiasen
  Lu, Devi Parikh, and Jason Weston. 2017{\natexlab{a}}.
\newblock {P}arl{AI}: A dialog research software platform.
\newblock In \emph{Proceedings of the 2017 Conference on Empirical Methods in
  Natural Language Processing: System Demonstrations}, pages 79--84,
  Copenhagen, Denmark. Association for Computational Linguistics.

\bibitem[{Miller et~al.(2017{\natexlab{b}})Miller, Feng, Batra, Bordes, Fisch,
  Lu, Parikh, and Weston}]{MillerFBBFLPW17}
Alexander~H. Miller, Will Feng, Dhruv Batra, Antoine Bordes, Adam Fisch, Jiasen
  Lu, Devi Parikh, and Jason Weston. 2017{\natexlab{b}}.
\newblock Parlai: {A} dialog research software platform.
\newblock In \emph{Proceedings of the 2017 Conference on Empirical Methods in
  Natural Language Processing, {EMNLP} 2017, Copenhagen, Denmark, September
  9-11, 2017 - System Demonstrations}, pages 79--84. Association for
  Computational Linguistics.

\bibitem[{Papangelis et~al.(2020)Papangelis, Namazifar, Khatri, Wang, Molino,
  and Tur}]{papangelis2020plato}
Alexandros Papangelis, Mahdi Namazifar, Chandra Khatri, Yi-Chia Wang, Piero
  Molino, and Gokhan Tur. 2020.
\newblock \href {http://arxiv.org/abs/2001.06463} {Plato dialogue system: A
  flexible conversational ai research platform}.

\bibitem[{Papineni et~al.(2002)Papineni, Roukos, Ward, and Zhu}]{PapineniRWZ02}
Kishore Papineni, Salim Roukos, Todd Ward, and Wei{-}Jing Zhu. 2002.
\newblock Bleu: a method for automatic evaluation of machine translation.
\newblock In \emph{Proceedings of the 40th Annual Meeting of the Association
  for Computational Linguistics, July 6-12, 2002, Philadelphia, PA, {USA}},
  pages 311--318. {ACL}.

\bibitem[{Paszke et~al.(2019)Paszke, Gross, Massa, Lerer, Bradbury, Chanan,
  Killeen, Lin, Gimelshein, Antiga, Desmaison, K{\"{o}}pf, Yang, DeVito,
  Raison, Tejani, Chilamkurthy, Steiner, Fang, Bai, and
  Chintala}]{PaszkeGMLBCKLGA19}
Adam Paszke, Sam Gross, Francisco Massa, Adam Lerer, James Bradbury, Gregory
  Chanan, Trevor Killeen, Zeming Lin, Natalia Gimelshein, Luca Antiga, Alban
  Desmaison, Andreas K{\"{o}}pf, Edward Yang, Zachary DeVito, Martin Raison,
  Alykhan Tejani, Sasank Chilamkurthy, Benoit Steiner, Lu~Fang, Junjie Bai, and
  Soumith Chintala. 2019.
\newblock Pytorch: An imperative style, high-performance deep learning library.
\newblock In \emph{Advances in Neural Information Processing Systems 32: Annual
  Conference on Neural Information Processing Systems 2019, NeurIPS 2019,
  December 8-14, 2019, Vancouver, BC, Canada}, pages 8024--8035.

\bibitem[{Qi et~al.(2020)Qi, Yan, Gong, Liu, Duan, Chen, Zhang, and
  Zhou}]{QiYGLDCZ020}
Weizhen Qi, Yu~Yan, Yeyun Gong, Dayiheng Liu, Nan Duan, Jiusheng Chen, Ruofei
  Zhang, and Ming Zhou. 2020.
\newblock Prophetnet: Predicting future n-gram for sequence-to-sequence
  pre-training.
\newblock In \emph{Proceedings of the 2020 Conference on Empirical Methods in
  Natural Language Processing: Findings, {EMNLP} 2020, Online Event, 16-20
  November 2020}, pages 2401--2410. Association for Computational Linguistics.

\bibitem[{Radford et~al.(2019)Radford, Wu, Child, Luan, Amodei, and
  Sutskever}]{radford2019language}
Alec Radford, Jeffrey Wu, Rewon Child, David Luan, Dario Amodei, and Ilya
  Sutskever. 2019.
\newblock Language models are unsupervised multitask learners.
\newblock \emph{OpenAI blog}, 1(8):9.

\bibitem[{Raffel et~al.(2020)Raffel, Shazeer, Roberts, Lee, Narang, Matena,
  Zhou, Li, and Liu}]{RaffelSRLNMZLL20}
Colin Raffel, Noam Shazeer, Adam Roberts, Katherine Lee, Sharan Narang, Michael
  Matena, Yanqi Zhou, Wei Li, and Peter~J. Liu. 2020.
\newblock Exploring the limits of transfer learning with a unified text-to-text
  transformer.
\newblock \emph{J. Mach. Learn. Res.}, 21:140:1--140:67.

\bibitem[{Rothe et~al.(2020)Rothe, Narayan, and Severyn}]{Rothe-TACL-2020}
Sascha Rothe, Shashi Narayan, and Aliaksei Severyn. 2020.
\newblock Leveraging pre-trained checkpoints for sequence generation tasks.
\newblock \emph{Trans. Assoc. Comput. Linguistics}, 8:264--280.

\bibitem[{See et~al.(2017)See, Liu, and Manning}]{SeeLM17}
Abigail See, Peter~J. Liu, and Christopher~D. Manning. 2017.
\newblock Get to the point: Summarization with pointer-generator networks.
\newblock In \emph{Proceedings of the 55th Annual Meeting of the Association
  for Computational Linguistics, {ACL} 2017, Vancouver, Canada, July 30 -
  August 4, Volume 1: Long Papers}, pages 1073--1083. Association for
  Computational Linguistics.

\bibitem[{Semeniuta et~al.(2017)Semeniuta, Severyn, and
  Barth}]{Semeniuta-EMNLP-2017}
Stanislau Semeniuta, Aliaksei Severyn, and Erhardt Barth. 2017.
\newblock A hybrid convolutional variational autoencoder for text generation.
\newblock In \emph{Proceedings of the 2017 Conference on Empirical Methods in
  Natural Language Processing, {EMNLP} 2017, Copenhagen, Denmark, September
  9-11, 2017}, pages 627--637.

\bibitem[{Serban et~al.(2016)Serban, Sordoni, Bengio, Courville, and
  Pineau}]{SerbanSBCP16}
Iulian~Vlad Serban, Alessandro Sordoni, Yoshua Bengio, Aaron~C. Courville, and
  Joelle Pineau. 2016.
\newblock Building end-to-end dialogue systems using generative hierarchical
  neural network models.
\newblock In \emph{Proceedings of the Thirtieth {AAAI} Conference on Artificial
  Intelligence, February 12-17, 2016, Phoenix, Arizona, {USA}}, pages
  3776--3784. {AAAI} Press.

\bibitem[{Sutskever et~al.(2014)Sutskever, Vinyals, and
  Le}]{Sutskever-NIPS-2014}
Ilya Sutskever, Oriol Vinyals, and Quoc~V. Le. 2014.
\newblock Sequence to sequence learning with neural networks.
\newblock In \emph{Advances in Neural Information Processing Systems 27: Annual
  Conference on Neural Information Processing Systems 2014, December 8-13 2014,
  Montreal, Quebec, Canada}, pages 3104--3112.

\bibitem[{Tang et~al.(2016)Tang, Yang, Carton, Zhang, and Mei}]{TangYCZM16}
Jian Tang, Yifan Yang, Samuel Carton, Ming Zhang, and Qiaozhu Mei. 2016.
\newblock \href {http://arxiv.org/abs/1611.09900} {Context-aware natural
  language generation with recurrent neural networks}.
\newblock \emph{CoRR}, abs/1611.09900.

\bibitem[{Vaswani et~al.(2018)Vaswani, Bengio, Brevdo, Chollet, Gomez, Gouws,
  Jones, Kaiser, Kalchbrenner, Parmar, Sepassi, Shazeer, and
  Uszkoreit}]{vaswani-etal-2018-tensor2tensor}
Ashish Vaswani, Samy Bengio, Eugene Brevdo, Francois Chollet, Aidan Gomez,
  Stephan Gouws, Llion Jones, {\L}ukasz Kaiser, Nal Kalchbrenner, Niki Parmar,
  Ryan Sepassi, Noam Shazeer, and Jakob Uszkoreit. 2018.
\newblock {T}ensor2{T}ensor for neural machine translation.
\newblock In \emph{Proceedings of the 13th Conference of the Association for
  Machine Translation in the {A}mericas (Volume 1: Research Track)}, pages
  193--199, Boston, MA. Association for Machine Translation in the Americas.

\bibitem[{Vaswani et~al.(2017)Vaswani, Shazeer, Parmar, Uszkoreit, Jones,
  Gomez, Kaiser, and Polosukhin}]{VaswaniSPUJGKP17}
Ashish Vaswani, Noam Shazeer, Niki Parmar, Jakob Uszkoreit, Llion Jones,
  Aidan~N. Gomez, Lukasz Kaiser, and Illia Polosukhin. 2017.
\newblock Attention is all you need.
\newblock In \emph{Advances in Neural Information Processing Systems 30: Annual
  Conference on Neural Information Processing Systems 2017, December 4-9, 2017,
  Long Beach, CA, {USA}}, pages 5998--6008.

\bibitem[{Vijayakumar et~al.(2018)Vijayakumar, Cogswell, Selvaraju, Sun, Lee,
  Crandall, and Batra}]{VijayakumarCSSL18}
Ashwin~K. Vijayakumar, Michael Cogswell, Ramprasaath~R. Selvaraju, Qing Sun,
  Stefan Lee, David~J. Crandall, and Dhruv Batra. 2018.
\newblock Diverse beam search for improved description of complex scenes.
\newblock In \emph{Proceedings of the Thirty-Second {AAAI} Conference on
  Artificial Intelligence, (AAAI-18), the 30th innovative Applications of
  Artificial Intelligence (IAAI-18), and the 8th {AAAI} Symposium on
  Educational Advances in Artificial Intelligence (EAAI-18), New Orleans,
  Louisiana, USA, February 2-7, 2018}, pages 7371--7379. {AAAI} Press.

\bibitem[{Yang et~al.(2019)Yang, Dai, Yang, Carbonell, Salakhutdinov, and
  Le}]{Yang-NIPS-2019}
Zhilin Yang, Zihang Dai, Yiming Yang, Jaime~G. Carbonell, Ruslan Salakhutdinov,
  and Quoc~V. Le. 2019.
\newblock Xlnet: Generalized autoregressive pretraining for language
  understanding.
\newblock In \emph{Advances in Neural Information Processing Systems 32: Annual
  Conference on Neural Information Processing Systems 2019, NeurIPS 2019,
  December 8-14, 2019, Vancouver, BC, Canada}, pages 5754--5764.

\bibitem[{Yang et~al.(2017)Yang, Hu, Salakhutdinov, and
  Berg{-}Kirkpatrick}]{Yang-ICML-2017}
Zichao Yang, Zhiting Hu, Ruslan Salakhutdinov, and Taylor Berg{-}Kirkpatrick.
  2017.
\newblock Improved variational autoencoders for text modeling using dilated
  convolutions.
\newblock In \emph{Proceedings of the 34th International Conference on Machine
  Learning, {ICML} 2017, Sydney, NSW, Australia, 6-11 August 2017}, pages
  3881--3890.

\bibitem[{Yu et~al.(2017)Yu, Zhang, Wang, and Yu}]{Yu-AAAI-2017}
Lantao Yu, Weinan Zhang, Jun Wang, and Yong Yu. 2017.
\newblock Seqgan: Sequence generative adversarial nets with policy gradient.
\newblock In \emph{Proceedings of the Thirty-First {AAAI} Conference on
  Artificial Intelligence, February 4-9, 2017, San Francisco, California,
  {USA}}, pages 2852--2858.

\bibitem[{Yu et~al.(2020)Yu, Zhu, Li, Hu, Wang, Ji, and Jiang}]{abs-2010-04389}
Wenhao Yu, Chenguang Zhu, Zaitang Li, Zhiting Hu, Qingyun Wang, Heng Ji, and
  Meng Jiang. 2020.
\newblock \href {http://arxiv.org/abs/2010.04389} {A survey of
  knowledge-enhanced text generation}.
\newblock \emph{CoRR}, abs/2010.04389.

\bibitem[{Zhang et~al.(2017)Zhang, Gan, Fan, Chen, Henao, Shen, and
  Carin}]{Zhang-ICML-2017}
Yizhe Zhang, Zhe Gan, Kai Fan, Zhi Chen, Ricardo Henao, Dinghan Shen, and
  Lawrence Carin. 2017.
\newblock Adversarial feature matching for text generation.
\newblock In \emph{Proceedings of the 34th International Conference on Machine
  Learning, {ICML} 2017, Sydney, NSW, Australia, 6-11 August 2017}, pages
  4006--4015.

\bibitem[{Zhu et~al.(2018)Zhu, Lu, Zheng, Guo, Zhang, Wang, and
  Yu}]{ZhuLZGZWY18}
Yaoming Zhu, Sidi Lu, Lei Zheng, Jiaxian Guo, Weinan Zhang, Jun Wang, and Yong
  Yu. 2018.
\newblock Texygen: {A} benchmarking platform for text generation models.
\newblock In \emph{The 41st International {ACM} {SIGIR} Conference on Research
  {\&} Development in Information Retrieval, {SIGIR} 2018, Ann Arbor, MI, USA,
  July 08-12, 2018}, pages 1097--1100. {ACM}.

\end{thebibliography}
\bibliographystyle{acl_natbib}

\end{document}